\documentclass[letterpaper, 10 pt, journal, twoside]{IEEEtran}
\IEEEoverridecommandlockouts                              % This command is only needed if 
\usepackage{gensymb}
\usepackage{cclicenses, graphicx}
\usepackage{amsmath}
\usepackage{array}
\usepackage{flushend}
\usepackage{balance}
\usepackage{multirow}
\usepackage[caption=false]{subfig}

\newcolumntype{C}[1]{>{\centering\arraybackslash}m{#1}}
\usepackage{tikz}
\usepackage{color}
\definecolor{darkgreen}{RGB}{0,127,0}
\definecolor{darkblue}{RGB}{0,0,175}
\newcommand{\addedtext}[2]{{{#2}}}
\newcommand{\modifiedtext}[2]{{{#2}}}
\newcommand{\deletedtext}[2]{{}}

\begin{document}
\title{User-Driven Functional Movement Training with a Wearable Hand Robot after Stroke}

\author{Sangwoo~Park$^{1}$, Michaela~Fraser$^{2}$, Lynne~M.~Weber$^{2}$, Cassie~Meeker$^{1}$, Lauri~Bishop$^{3}$, Daniel~Geller$^{2}$, Joel~Stein$^{2,4}$, and Matei~Ciocarlie$^{1,4}$%
\thanks{This work was supported in part by the National Science Foundation under grant IIS-1526960 (part of the National Robotics Initiative).}%
\thanks{$^{1}$Department of Mechanical Engineering, Columbia University, New York, NY 10027, USA.}%
\thanks{\hspace{-3mm}{\tt\footnotesize \{sp3287, cgm2144, matei.ciocarlie\}@columbia.edu}}%
\thanks{$^{2}$Department of Rehabilitation and Regenerative Medicine, Columbia University, New York, NY 10032, USA. {\tt\footnotesize \{mgf2124, lw2739, dlg34, js1165\}@cumc.columbia.edu}}%
\thanks{$^{3}$Division of Biokinesiology and Physical Therapy, University of Southern California, Los Angeles, CA 90089, USA. {\tt\footnotesize lauribis@usc.edu}}
\thanks{$^{4}$Co-Principal Investigators}
}

\markboth{IEEE TRANSACTIONS ON NEURAL SYSTEMS AND REHABILITATION ENGINEERING}%
{Park \MakeLowercase{\textit{et al.}}: User-Driven Functional Movement Training with a Wearable Hand Robot after Stroke}

\maketitle

%%%%%%%%%%%%%%%%%%%%%%%%%%%%%%%%%%%%%%%%%%%%%%%%%%%%%%%%%%%%%%%%%%%%%%%%%%%%%%%%
\begin{abstract}
We studied the performance of a robotic orthosis designed to assist the paretic hand after stroke. It is wearable and fully user-controlled, serving two possible roles: as a therapeutic tool that facilitates device-mediated hand exercises to recover neuromuscular function or as an assistive device for use in everyday activities to aid functional use of the hand. We present the clinical outcomes of a pilot study designed as a feasibility test for these hypotheses. 11 chronic stroke ($>$ 2 years) patients with moderate muscle tone (Modified Ashworth Scale $\leq$ 2 in upper extremity) engaged in a month-long training protocol using the orthosis. Individuals were evaluated using standardized outcome measures, both with and without orthosis assistance. Fugl-Meyer post intervention scores without robotic assistance showed improvement focused specifically at the distal joints of the upper limb, suggesting the use of the orthosis as a rehabilitative device for the hand. Action Research Arm Test scores post intervention with robotic assistance showed that the device may serve an assistive role in grasping tasks. These results highlight the potential for wearable and user-driven robotic hand orthoses to extend the use and training of the affected upper limb after stroke.

\end{abstract}

%%%%%%%%%%%%%%%%%%%%%%%%%%%%%%%%%%%%%%%%%%%%%%%%%%%%%%%%%%%%%%%%%%%%%%%%%%%%%%%%

\section{Introduction}

\IEEEPARstart{H}{emiparesis} of the upper limb (UL) is a common and debilitating complication after stroke~\cite{LAWRENCE01}. Approximately 50\% of survivors with UL paralysis continue to present with functional deficits four years after stroke~\cite{g1999long}. Growing evidence demonstrates high quality, highly repetitive, and task-specific training is beneficial in UL recovery after stroke~\cite{lang2016dose,schneider2016increasing}. However, there are challenges which impede many chronic stroke patients from receiving this type of rehabilitation program, for reasons that include logistical and geographical barriers of visiting therapy clinics, insurance and reimbursement limitations, and insufficient availability of therapists with specialized training~\cite{shortage2014}.

\begin{figure}[t]
	\centering
	\vspace{-3mm}
	%	\subfloat[Exotendon device and EMG armband.]{%
	\includegraphics[width=1\linewidth]{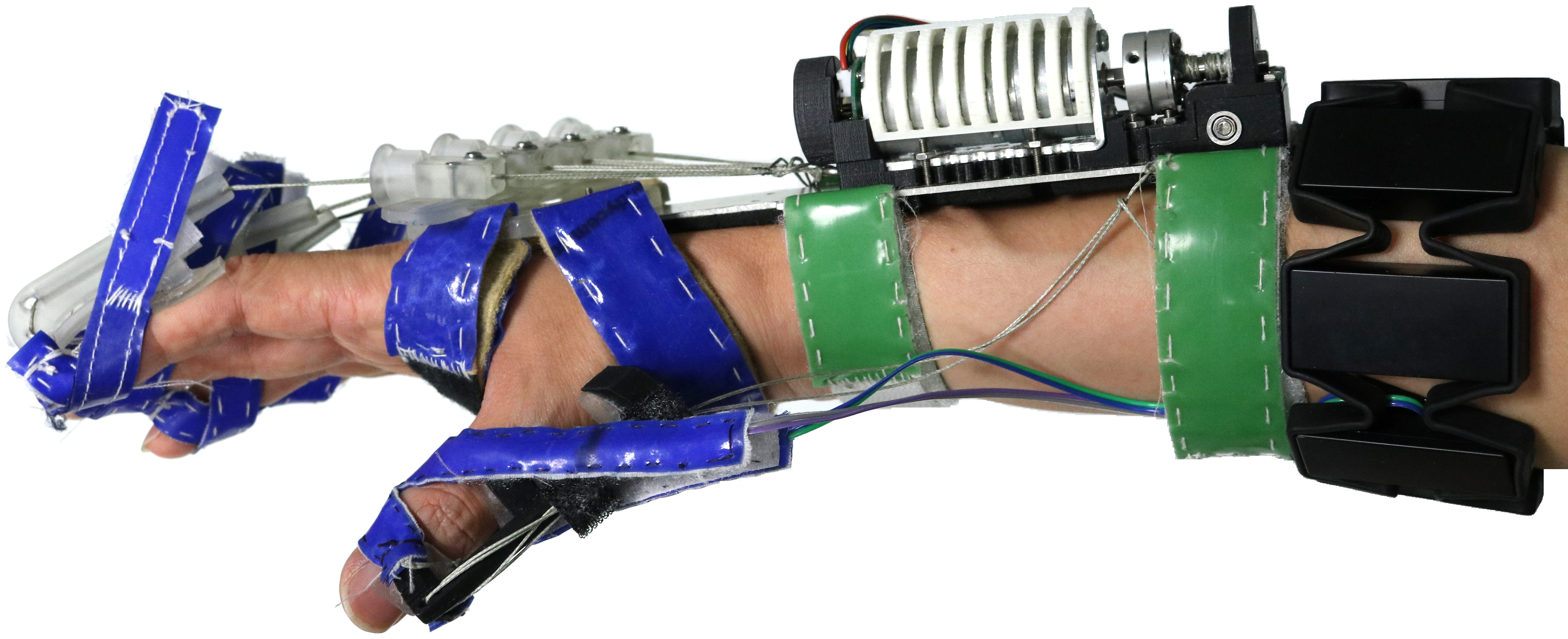}
	\vspace{-2mm}
	
	\includegraphics[width=1\linewidth]{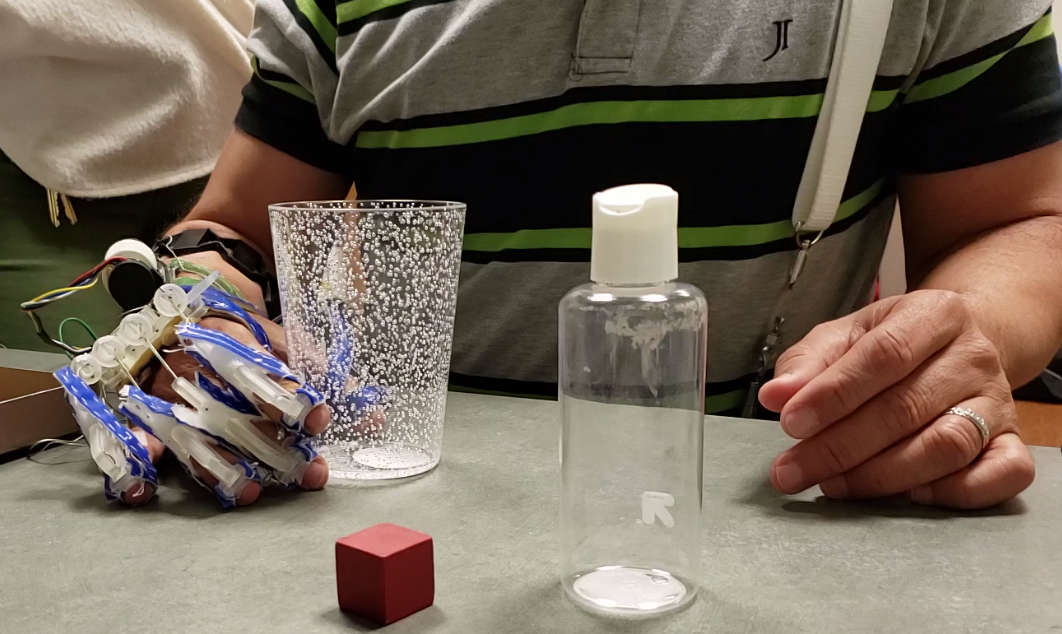}
	\caption{Top : Exotendon device and EMG armband. Bottom : Functional movement training with a wearable hand orthosis.}
	\label{fig:exotendon_training}
	\vspace{-5mm}
	%	}
\end{figure}

%There is a general consensus in the field that additional repetitions in therapy would be highly beneficial to numerous patient populations~\cite{LANG07}. However, while occupational therapy is effective for hand rehabilitation, insufficient therapy is available to stroke survivors, for reasons that include the logistical burdens of visiting a clinical facility, reimbursement limitations, as well as a shortage of available therapists~\cite{shortage2014}.

Robotic devices have been designed to address the need for increased UL repetitions for post-stroke rehabilitation~\cite{krebs2018twenty+}. However, these largely target proximal segments of the UL (i.e. shoulder and elbow); in studies with these devices, Fugl-Meyer (FM) - UL and kinematic parameters show that robotic training produces improved functional gains mostly in the proximal joints~\cite{lum2002robot,calabro2016may}. This disparity of motor recovery between proximal and distal joints (i.e. wrist and fingers) can lead to undesirable compensatory grasp patterns, which make long-term rehabilitation more complicated and often ineffective~\cite{gillen2015stroke}. In contrast, recent work found clinical effectiveness when subjects receive robotic support directly to the distal joints compared to support to the proximal joints alone~\cite{qiuyang2019distal}. However, wearable active assistance for the hand is complicated by the limited space available for the necessary motor, controller, transmission components, and power storage, as well as by the complex anatomy and kinesiology of the stroke-affected hand. These difficulties have led to the current lack of functional prototypes for wearable hand robots used and assessed with clinical efficacy~\cite{lambercy2018robot}.

To help address this need, we have developed a wearable hand orthosis, which provides mechanical assistance for finger extension. Our previous studies have established the basic operation principles of the device: in limited case series with stroke survivors, we have shown that our exotendon network can facilitate finger extension to enable gross grasp/release~\cite{park2016on} via a small motor relying on effective force transmission mechanisms~\cite{park2018design}, and that, for a subset of patients, we can infer the intent to open and close the hand when the orthosis is used in conjunction with a commodity electromyography (EMG) armband~\cite{meeker2017emg}. However, the clinical performance of this device and the importance of training effects over longer-term use have not been investigated to date. 

In this pilot study, we focus on the performance of our hand orthosis as a \textit{rehabilitation device} and as an \textit{assistive device}. We present clinical outcomes from a 12 session training program, comprising 3 sessions per week for 4 weeks. Each session involved 30 minutes of active training time in which participants practiced a variety of grasp and release tasks with everyday objects to simulate Activities of Daily Living (ADLs). 11 subjects with chronic stroke completed the protocol and were evaluated with a battery of clinical assessments pre- and post-intervention: FM - UL,  Action Research Arm Test (ARAT), and Box and Block Test (BBT).

In order to determine the efficacy of the orthosis as a \textit{rehabilitative device}, we compare clinical outcomes both pre- and post-intervention \textit{without device assistance}. To distinguish between recovery throughout the entire UL and more localized improvement of the distal segments, we subdivide FM into shoulder/elbow (FM-proximal) and wrist/hand (FM-distal). To determine efficacy when used as an \textit{assistive device}, we compare baseline performance and post-intervention clinical outcomes without assistance to post-intervention performance \textit{while wearing the device}. Our goal is to determine if increased competence using the device from a month-long training program leads to increased performance in tasks requiring grasp, transport, and release of objects while wearing the orthosis. We examine the assistive capability through ARAT scores measured while the users are wearing the robot. Finally, we provide secondary analyses to compare performance for our two intent inferral methods, namely EMG and shoulder harness controls, used to drive the hand device during both training and post-testing.

Overall, the main contributions of this paper are as follows:
\begin{itemize}
	\item To the best of our knowledge, it is the first time an active wearable hand robot was evaluated in clinical assessments both with and without robotic assistance, following user-driven functional hand exercises over multiple training sessions for chronic stroke patients interacting with various real-world objects.
	\item FM subsection scores suggest that intensive hand functional exercises using our robot may improve motor function in distal segments on the UL.
	\item Subsection of ARAT highlight the potential for the assistive capability of the device for grasping components of ADLs.
\end{itemize}
\noindent Our aim is to provide wearable, user-driven assistance for the stroke-affected hand, and to investigate its effects on both rehabilitation and functional performance. These are steps towards our long-term goal to test user-operated active orthoses outside of a clinical setting, for rehabilitation or home-assistance, with the potential to significantly increase the number of motor repetitions and the intensity of training.

\section{Related Work}
%In the past, the main focus of numerous robotic UL rehabilitation devices has been on the shoulder and elbow. This type of device provides repetitive UL exercises through large exoskeletal workstations~\cite{lo2010robot,krebs2007robot}. Even though studies with these devices offer some support that a robotic rehabilitative treatment can be effective in motor recovery of the UL, improvement is only reported for proximal segments~\cite{lum2002robot,calabro2016may}. Therefore these subjects may still experience functional limitations as it is difficult to complete most ADLs without improvement in the UL distal segments. 

Due to the complexity of hand movements and variable impairment patterns seen in stroke patients, it is only recently that robotic devices for hand rehabilitation have been proposed~\cite{lambercy2018robot}. Robotic workstations for hand rehabilitation on the market have demonstrated their feasibility and efficacy~\cite{vanoglio2017feasibility,susanto2015efficacy}. But, workstation devices are often bulky, costly, and tethered to a clinical setup requiring extensive supervision by health care providers. Wearable robots, by contrast, have potential to enable use outside the clinic allowing a greater number of repetitions of functional tasks. Therefore, this type of robot has become a focus of recent research in robot-assisted rehabilitation~\cite{lang2007counting}. To facilitate these benefits, we focus on wearable hand devices in this work, though in this early pilot stage, the device is utilized in a clinical environment with staff supervision.

Traditional wearable hand robots in the early stage are mostly exoskeleton comprised of rigid links~\cite{heo2012current}. However, such devices are difficult to align axes of biological and robotic joints. A soft pneumatic actuator based orthosis can be an alternative as it provides safe human-robotic interaction due to natural compliance and flexibility. Polygerinos et al. have developed a pneumatic powered glove that is inexpensive, low-profile, and well adapted to complex finger movement~\cite{polygerinos2015soft}, although untethering from air pressure sources can be challenging for complete portability. A wearable hand device can also take the form of supernumerary robotic finger to provide assistive benefits for practical use of the affected UL~\cite{hussain2016soft}.

For intuitive and user-driven control of a wearable hand device, there has been research to develop wearable sensors. Multichannel EMG with wireless communication can be built within a wearable package~\cite{brunelli2015low}, and this type of sensor allows intuitive control using a pattern recognition algorithm~\cite{meeker2017emg},~\cite{ryser2017fully}. A low-profile and portable glove with various sensors, such as DAGLOVE~\cite{weber2016low}, can also allow use of pattern recognition utilizing multimodal sensor data. Multimodal intent detection can provide better accuracy for stroke patients compared to an algorithm that uses a single modality, such as EMG~\cite{park2018multimodal}.

Among many hand rehabilitation tools~\cite{bos2016structured,lambercy2018robot}, a subset have been supported by clinical evidence for chronic stroke patients. The PneuGlove, which facilitates manipulation activities in virtual reality and with real-world objects, demonstrated improvements in a pilot study of 14 subjects~\cite{connelly2010pneumatic}. Hand of Hope has also proven the efficacy of robot-assisted training through a pilot study with 19 stroke patients~\cite{susanto2015efficacy}. A number of passive devices have also been proposed and validated~\cite{ates2017script,lannin2016upper}, with the advantage of being simpler to operate for unsupervised therapy. However, given that patients with hemiparesis may exhibit severe muscle weakness in grip strength~\cite{boissy1999maximal}, even lower level of mechanical interference with finger flexion via passive mechanisms can adversely affect hand functionality.

While the majority of devices aim to demonstrate clinical effectiveness as a rehabilitative training tool, some others aim to develop an assistive device for immediate assisted functional gains. Yurkewich et al. have shown that three chronic and two acute stroke subjects improved in modified BBT, range of motion of the fingers, and a subset of CAHAI~\cite{yurkewich2019hand}. The other device is the Myomo which is a portable elbow-wrist-hand orthosis controlled by EMG. With the Myomo, 18 chronic stroke survivors achieved immediate and significant UL improvements as confirmed by FM, BBT, and a battery of functional tasks~\cite{peters2017giving}. In terms of intent detection for assistive robots, a mathematical formulation and algorithm can potentially help improve functionality of users with motor impairment~\cite{jain2018recursive}. However, these devices have not been tested for long-term use for ADLs. We believe it is important for wearable devices to demonstrate their robustness such that users can complete clinical sessions, and eventually use the device unsupervised at home. In this work, we investigate the feasibility of our hand robot through a clinical protocol to evaluate its strengths and limitations, both as a rehabilitative intervention and as an assistive device.

\section{An Active Hand Orthosis for Stroke Patients}

Chronic stroke patients with hemiparesis often experience functional disuse of their hand. Our device aims to address one of the most common impairment patterns seen in this population: a combination of weakness, spasticity, poor coordination, and a flexor synergy pattern where individuals may be able to actively flex their fingers to form a gross grasp, but are unable to volitionally extend their fingers to release the grasp. By assisting finger extension, our device enables users to harness their residual function and incorporate their impaired hand into	functional grasp and release tasks.

\begin{figure}[t]
	\centering
	\includegraphics[width=1\linewidth]{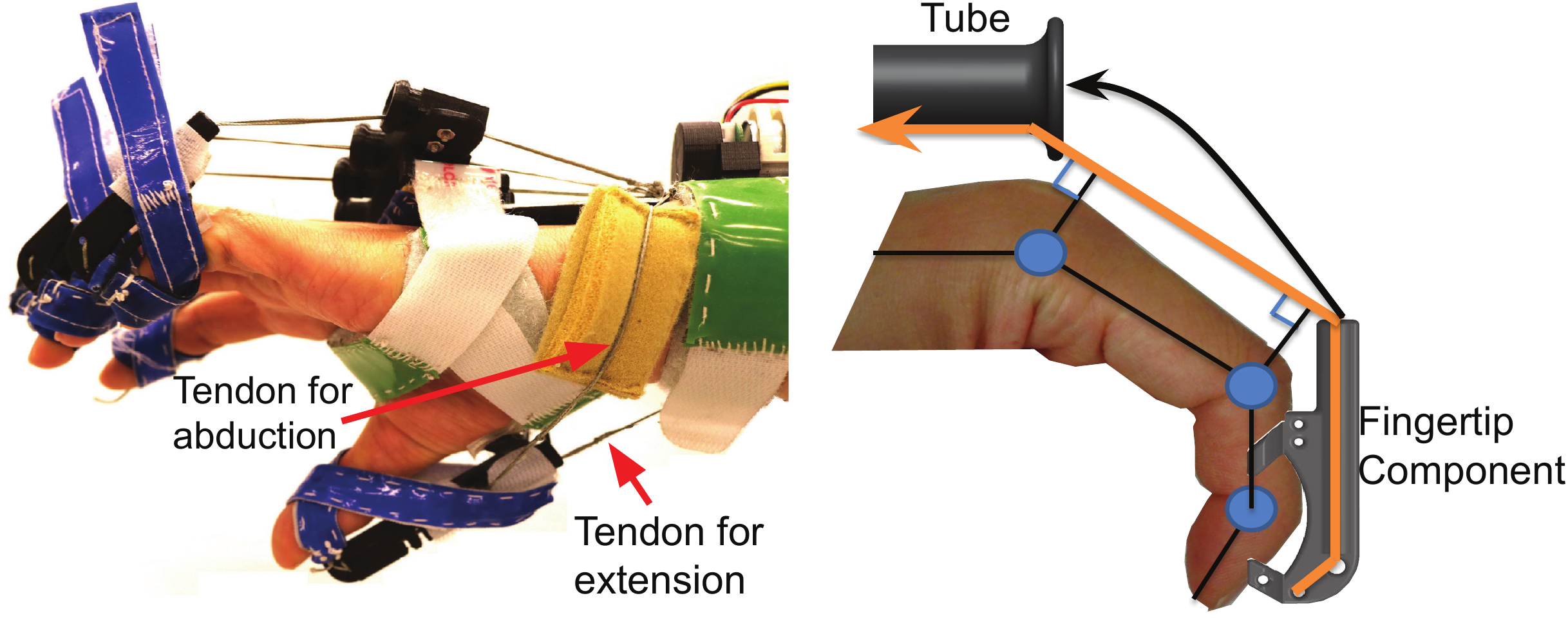}
	\vspace{-7mm}
	\caption{Tendon routes for the thumb (left) and fingertip components (right).}
	\label{fig:fingertip_components}
\end{figure}

\subsection{Exotendon Device}

To address many practical and anatomical challenges of hand robot designs, our device utilizes underactuation mechanisms through tendon networks~\cite{park2016on}. The anchoring structures are designed to provide effective force transmission, in order to overcome hand spasticity while using low motor forces~\cite{park2018design}. This design allows functional hand movement using a small actuator while also reducing distal migration, an undesirable phenomenon where the motor components of the device slide down the forearm towards the hand due to the applied forces.

Our exotendon device (Fig.~\ref{fig:exotendon_training} - top) consists of a rigid forearm splint on which an actuator (Pololu-25D-MP-12V) is mounted, along with 3D printed distal components for the fingers. The splint constrains wrist movement at neutral angle, thus allowing motor force to be transmitted through cable routes to the fingers. The 3D printed fingertip components (Fig.~\ref{fig:fingertip_components} - right) are secured on the dorsal side with Velcro strapped around each finger. The main role of these components is to increase the tendon moment arms around the proximal interphalangeal (PIP) joints throughout the entire range of motion, so that the motor force is effectively applied to the digits. Also, this component mechanically prevents hyper-extension of the distal interphalangeal (DIP) joint to avoid any injury. Similarly, the distal portion of the fingertip component creates a mechanical block that prevents hyper-extension at the PIP and metacarpophalangeal (MCP) joints. \addedtext{AE-1}{The device does not prevent hyper-extension of the PIP joints for stroke patients with significantly higher stiffness on the MCP joints than on the PIP joints; however, in our study we have not encountered any patients exhibiting this pattern.}

The main role of the orthosis is to provide assistance for finger extension. As size and weight are of critical importance for building a wearable device, we rely on a heavily underactuated design: a single motor provides assistance for extending all the digits except the thumb. When we detect the wearer's intent to open the hand (as detailed in the next subsection), the actuator retracts, applying extension torques to the digit joints via the exotendon network. Conversely, when we detect the intent to close the hand, the motor extends and relaxes the forces in the tendon network, allowing the digits to flex. We rely on the patient being able to generate sufficient finger flexion torques, which is common for the impairment pattern we focus on. Throughout this study, we use a PID position control to drive the motor, and the range of motion is determined at the beginning of protocol depending on the size of the user's hand. Additional specifications include the following - weight: 365 g; motor gear ratio: 47:1; extension/retraction time: 1.8 s; max extension force: 100 N; donning and doffing time: approx. 15 mins and 1 min respectively.

%See Table~\ref{tab:device_specification} for more device specification.

The natural movements of the thumb are complex and distinct from the other digits~\cite{CUEVAS2003}. In addition to flexion/extension and abduction/adduction, the thumb moves in opposition to the other digits to grasp and pinch. To ensure that the thumb approximately opposes the other fingers, we use two passive cable routes, one for extension and another for abduction (Fig.~\ref{fig:fingertip_components} - left). The tensions of these two cables are calibrated at the beginning of each session, and are fixed for the duration of the session. The thumb is thus not actively assisted by the orthosis, but statically splinted into opposition for the duration of the protocol, as seen previously in the literature~\cite{arata2013}. We have found this procedure to enable grasping while reducing the load on the actuator, which must only assist the remaining digits.% For future designs, we plan to also investigate the addition of one or more actuators in order to provide independent thumb actuation.

%% \begin{table}
%% 	\caption{Device specification}
%% 	\centering
%% 	\label{tab:device_specification}
%% 	\begin{tabular}{l l}
%% %		& Back hand+fingertip components & 59g \\ 
%% %		& Splint+motor & 211g \\ 
%% %		& EMG armband & 95g \\ 
%% 		Weight & 365 g \\ \hline 
%% 		Motor gear ratio &  47:1 \\ \hline		
%% 		Extension/retraction time & 1.8 sec \\ \hline %$\approx$ 
%% 		Maximum extension force & 100 N \\ \hline
%% 		Donning time & 15 mins \\ \hline
%% 		Doffing time & 1 min
%% 	\end{tabular}
%% 	\vspace{-7.2mm}
%% \end{table}

\subsection{Intent Detection}

Our goal for this robotic device is for the patient to initiate robotic assistance by signaling the intent to open the hand (when the motor retracts, providing assistance for finger extension), or to close (when the motor extends, allowing the fingers to flex). We provide two methods for detecting the intent of the patient, and compare them here. 

\subsubsection{Intent Detection via Ipsilateral EMG Signals}

The first method utilizes ipsilateral forearm surface EMG signals to detect the user intent as the patient attempts to use the affected hand. If it can be realized, it has appealing characteristics: it is highly intuitive (the patient simply attempts to open/close the hand as needed for the task) and can facilitate neuroplasticty as it closes sensorimotor loop on the impaired UL.

In this study, we use an EMG-based intent inferral method introduced previously~\cite{meeker2017emg, park2018multimodal}. This approach relies on a pattern recognition algorithm to detect user intention based on data collected by commercial armband (Myo by Thalmic Labs) equipped with eight sensors, and does not require precise sensor placement. The armband is placed approximately one inch proximal to the splint to avoid contact with the motor.

While previous work has shown that intuitive control is indeed possible, it has also highlighted a number of challenges. EMG signals are inherently abnormal in patients with hemiparesis and can be distorted by spasticity and fatigue~\cite{cesqui2013}. As a result, when working with stroke patients while engaged in functional tasks, we found our current EMG-based intent detection method to be effective only for a subset of patients.

\begin{figure}[t]
	\centering
	\includegraphics[width=1\linewidth]{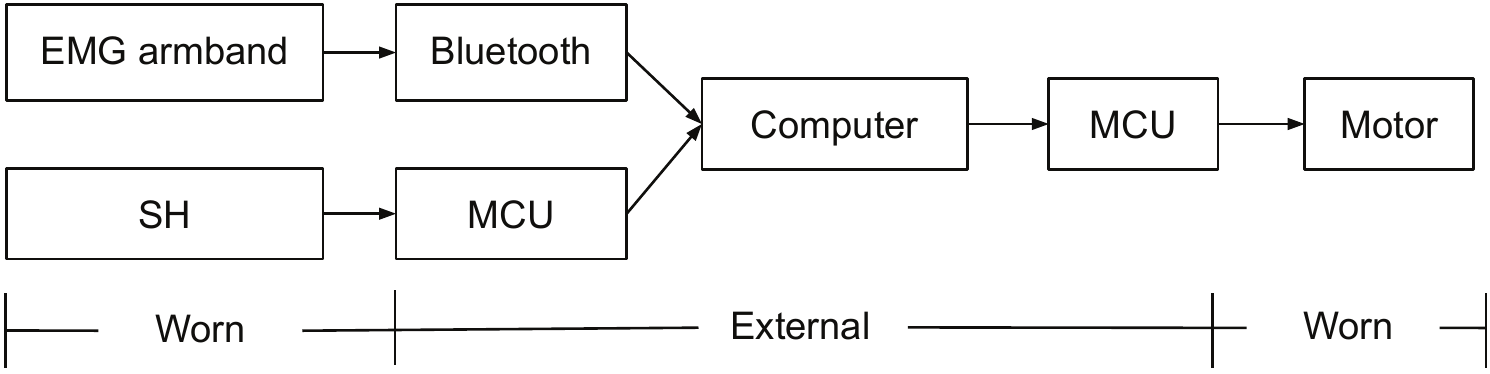}
	\caption{EMG armband or SH, depending on the assigned group, sends biophysical data to a computer through bluetooth or microcontroller unit (MCU). In the computer, intent detection algorithm classifies the intention and generate a motor command. Then, the command is transmitted to the motor through the MCU.}
	\label{fig:data_flow}
	\vspace{-2mm}
\end{figure}

\subsubsection{Intent Detection via Contralateral Shoulder Movement}

To account for this phenomenon, we introduce here a second intent inferral method, using contralateral shoulder movement. This approach, often used for body powered UL prostheses~\cite{van1983field}, provides a more robust control compared to EMG, as it relies on the unaffacted side. Additionally, compared to other non-EMG control methods, such as a button switch, it enables bimanual tasks since the unaffected hand is not required to operate the control. Thus, the focus with this method was on restoration of functionality, rather than neuro-recovery. It has the disadvantage of requiring the patient to engage in additional movement (elevating the contralateral shoulder) with the only purpose of providing a signal for our device. Such movement can be unintuitive, and of limited rehabilitative value. However, it may follow that improvements in functional use result in improved proximal strength and bilateral integration.  

\begin{figure}
	\centering
	\begin{minipage}{0.5\linewidth}
		\centering
		\includegraphics[width=1\linewidth]{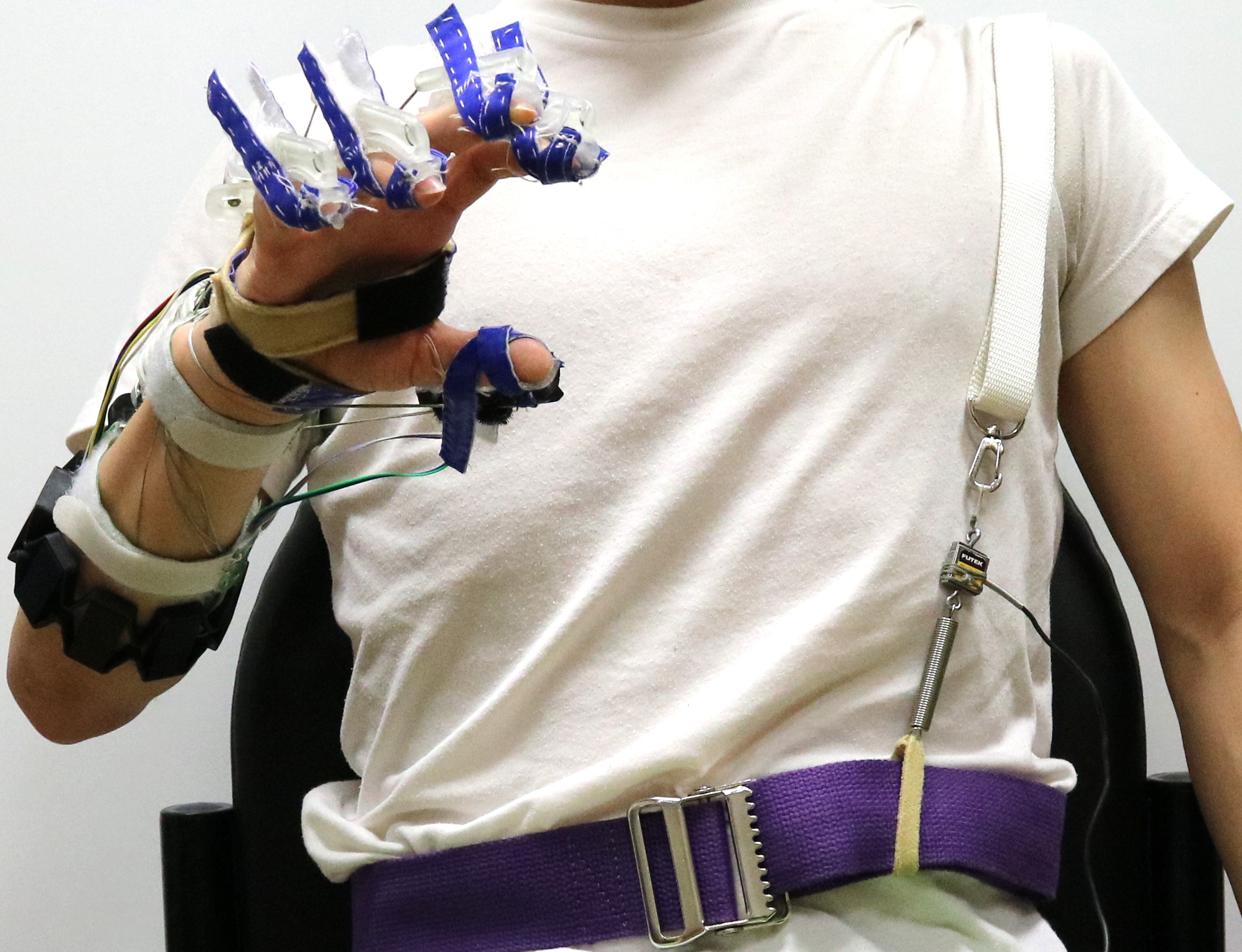}
		\vspace{-5mm}
		\caption{Exotendon device with SH method.}
		\label{fig:SH}
	\end{minipage}\hfill
	\begin{minipage}{.47\linewidth}
		\centering
		\includegraphics[width=1\linewidth]{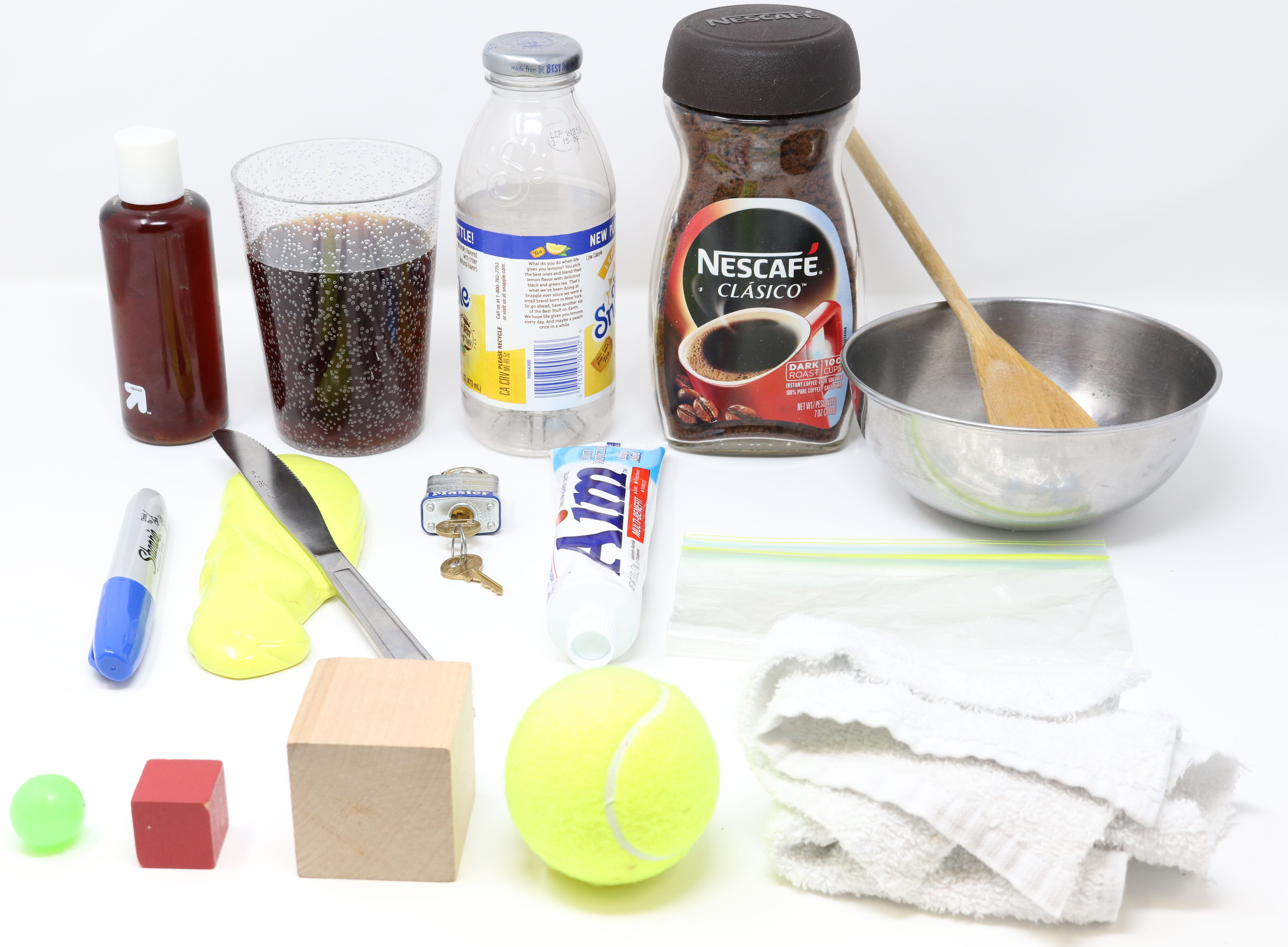}
		\vspace{-3mm}
		\caption{Real world objects used for treatment.}
		\label{fig:objects}
	\end{minipage}
\vspace{-1mm}
\end{figure}

We use a shoulder harness worn on the unimpaired UL and used to detect shoulder movement. When shoulder depression is detected, the device retracts to trigger hand opening through finger extension (Fig.~\ref{fig:SH}). Shoulder flexion was ruled out to control release, as it promoted a flexor synergy in the affected limb, while shoulder depression (often coupled with relaxation/exhalation) appeared more favorable to facilitate release. Conversely, when shoulder elevation (shrug) is detected, the device extends to allow hand closure via finger flexion.

A load cell (Futek, FSH00097) is installed in series with a suspender to measure the tension in the harness, and an extension spring next to the load cell connects a waist belt and the suspender to ease discomfort caused by the tension. Two different thresholds on the load cell signal are used to detect shoulder elevation and depression, in order to prevent unnecessary motor oscillation. The thresholds are manually calibrated at the beginning of each session. In the rest of this study, we will refer to this intent inferral method as SH, shorthand for shoulder harness.

The main advantage of the contralateral SH intent inferral method over EMG is its robustness to differences in impairment patterns, since it relies exclusively on the unimpaired side. Still, we believe that the potential advantages of ipsilateral EMG control (more intuitive motor commands, closing the loop on the affected side) outweigh the SH robustness advantage, as long as EMG control is applicable. %In consequence, we chose not to randomly assign intent detection methods for clinical intervention. Instead, users are assigned to use EMG method as long as it can correctly detect intention during a control screening, which we describe in section~\ref{subsec:protocol}. Otherwise, users are assigned to the SH group. 

%
%
%\begin{figure}
%	\centering
%	\begin{subfigure}{width=0.5\textwidth}
%		\centering
%		\includegraphics[width=0.45\linewidth]{images/SH.jpg}
%		\caption{Exotendon device with SH method.}
%		\label{fig:SH}
%	\end{subfigure}
%	\includegraphics[width=0.45\linewidth]{images/objects.jpg}
%	\caption{Real world objects used for treatment.}
%	\label{fig:object}	
%\end{figure}

\section{Clinical Intervention}
\label{sec:clinical_intervention}

In order to evaluate our active hand orthosis, we performed a clinical study aiming to quantify its performance as either a rehabilitative or assistive device. The three main characteristics of the study included the following: first, each training session consisted of user-controlled functional interaction with everyday objects and simulated ADLs. This is intended to emulate use of the impaired UL outside of clinical settings, which is our directional goal for the project. Second, each patient underwent twelve 30 minutes training sessions, distributed over the course of one month. The relatively large number of sessions (compared to our previous feasibility studies) was required both to study rehabilitative effects, and to allow patients to develop familiarity with the device and its controls, in order to quantify performance as an assistive device. Third, our outcome measures post-intervention included clinical assessments performed both assisted (with the device on, in order to study assistive performance) and unassisted (without the device, in order to study rehabilitation effects). Post-intervention clinical outcomes without robotic assistance were compared to baseline to evaluate rehabilitative effect while post-testing results with robotic assistance were compared to baseline performance and post-testing results without assistance to examine assistive capability of the device. We present the details of our clinical intervention next.

%\begin{figure}[t]
%	\centering
%	\includegraphics[width=0.55\linewidth]{images/objects.jpg}
%%	\subfloat[Objects in a cake pan]{%
%%	\includegraphics[width=0.45\linewidth]{images/pan_objects.jpg}\
%%	}
%	\caption{Real world objects used for treatment.}
%		\label{fig:objects}
%\end{figure}

\subsection{Participants}

Total twelve community-dwelling individuals with chronic stroke volunteered to participate in the study and met inclusion criteria. Inclusion criteria were:

\begin{itemize}
	\item Stroke diagnosis at least 6 months prior to start of study
	\item Passive range of motion: wrist to neutral, digits within normal limits
	\item Moderate muscle tone, i.e., Modified Ashworth Scale (MAS) $\leq$ 2 in digits, wrist, and elbow
	\item Active range of motion: At least 30 degrees shoulder flexion, 20 degrees shoulder abduction, 20 degrees elbow flexion, finger flexion within functional limits
	\item Strength: At least trace palpable finger extension
	\item Able to successfully flex the fingers to form a grasp
	\item Unable to extend all fingers fully without assistance
	\item Intact cognition to provide informed written consent
\end{itemize}

Exclusion criteria were:
\begin{itemize}
	\item Concurrent participation in another study
	\item Comorbid orthopedic condition/pain limiting functional use of the impaired upper extremity
	\item History or neurological disorder other than stroke
	\item Excessive spasticity (MAS $>$ 2)
	\item Recent botox injection to the affected limb ($<$ 13 weeks)
\end{itemize}

5 participants had prior experience with the exotendon device in varying capacities, but not within 6 months before start of the protocol. All subjects gave informed written consent to participate and the protocol was approved by the Columbia University Medical Center Institutional Review Board. Participants were primarily recruited through a pre-existing, IRB-approved, research registry of stroke patients. Additionally, physiatrists in our clinic referred some of their patients. The trial was registered on ClinicalTrials.gov (NCT03767894). All training and testing sessions were performed under the supervision of an occupational or physical therapist.

\subsection{Outcome Measures}

\subsubsection{Baseline Assessments}

All clinical assessments were performed by an occupational therapist who was not involved in the training protocol, though blinding was not possible in this study design. For all testing sessions, the MAS was performed first since other measurement tools can cause fatigue. After the MAS, the FM, ARAT, and BBT, were administered in a randomized order to limit order effect. FM for UL is an impairment level measure of body structures that evaluates reflexes, motor function, and joint range of motion of the UL~\cite{fugl1975post}, ARAT is an activity level assessment that involves specific grasp, grip, pinch, and gross motor tasks for the UL~\cite{lyle1981performance}, and BBT is an activity level assessment that tests unilateral pinch and manual dexterity in a timed manner~\cite{mathiowetz1985adult}. The maximum score on the FM for UL is 66, and the FM can be subscaled into FM-proximal (42/66), and FM-distal (24/66). Its estimated Minimal Clinically Important Difference (MCID) ranges from 4.25 to 7.25~\cite{page2012clinically}. The maximum score on the ARAT is 57, and anchor-based MCID of the ARAT for chronic stroke is 5.7~\cite{van2001intra}.

\subsubsection{Post Assessments}

To evaluate rehabilitative and assistive effects of the training, participants completed post-testing assessments, split over the course of two sessions to avoid fatigue. One session involved administration of the FM, ARAT, and BBT without robotic assistance, while the other session involved administration of the ARAT and BBT with robotic assistance. The order of post-testing days was also randomized. The FM was only performed at post-testing without robotic assistance because the FM assesses capacity of the arm primarily through gross motor tasks, and comparatively few grasping and pinching tasks. We thus presumed that robotic assistance would have minimal influence on FM scores.

%While our training protocol contains pick and place exercises using various real-world objects, which encourages functional and practical movements that are useful for ADLs, the number of motor repetitions is lower compared to other studies~\cite{lambercy2018robot}. Therefore, the motor gain from the outcome measures might be limited.

Post-testing without the device assesses motor recovery after robot-assisted training whereas post-testing with robotic assistance evaluates the assistive aspects of the device. We assumed that the proposed intent detection methods, particularly EMG-based, will take time for users to learn, so the clinical assessments are performed after 12 training sessions in order to allow competent use of the device. 

\subsubsection{Statistical Analysis}
Our primary outcome analyses include the FM, ARAT, and their subscales rated on all subjects. The FM has two sub-scales: FM-proximal evaluates the shoulder and elbow, while FM-distal evaluates the wrist and hand. The ARAT has 4 sub-scales: grasp, grip, pinch, and gross movement. Secondary outcome analyses include the FM and ARAT broken down by control method (EMG and SH), and BBT broken down by level of functionality of the participants. For all data, we report mean gains of unassisted post-training score compared to baseline score to examine the rehabilitative effect as well as assisted post-training score compared to baseline score and unassisted post-training score to assess the assistive effect.

We tested the primary clinical outcome data and gains for normal distribution based on the residuals of our dependent variables with Shapiro-Wilk and with Q-Q plots. We also tested the homogeneity of variance using the Levene test. FM and its subsections, ARAT, ARAT-grasp, and ARAT-grip passed the normality and homogeneity test, and we applied paired sample t-tests. ARAT-pinch, ARAT-gross movement failed the normality test, and we applied a nonparametric paired sample Wilcoxon test which does not make assumptions of normality or homogeneity of variance.

\addedtext{AE-3}{To determine statistical significance, we computed p-values for all of our primary measures. We utilized the Benjamini-Hochberg (BH) procedure to control the False Discovery Rate (FDR) for mutiple comparisons~\cite{benjamini1995controlling}, using 0.05 as the overall FDR control level. We note that, when using the BH procedure, the threshold used for rejecting the null hypothesis is different for each test, even if the overall control level is the same. We thus report, for every test, both the p-value and the BH significance threshold for rejecting the null hypothesis. Our secondary measures are considered post-hoc analyses, and we report only mean gains with no statistical analysis.}

\subsection{Protocol}
\label{subsec:protocol}

After participating in the informed consent process, all participants were screened for inclusion, and those included performed baseline measurements. During the next screening visit, each participant was fitted with the exotendon device to avoid hyper-extension and was screened with the EMG classifier to determine which control method they would use for the study. During the control screening, a classifier was trained with the subject's EMG signals, and the user was instructed on how to use the control. The user was asked to open, relax, and close their hand three times each, with the forearm on and off the table. If our EMG method was able to classify user intention correctly under all conditions, and the user could maintain each signal for at least 2 seconds on every attempt, the user was assigned to the EMG group. Otherwise, the participant was assigned to the SH group.

%We allowed the use of a mobile arm support for participants who had difficulty performing the functional tasks due to proximal weakness and/or significant fatigue. Criteria for use of mobile arm support include absent or weak (0 to fair grades) elbow flexion/extension, shoulder flexion and abduction, or external rotation~\cite{atkins2008mobile}. The participants that met these criteria used the arm support for the duration of training sessions.

We allowed the use of a mobile arm support (Saebo MAS) for participants who were clinically observed to have significant difficulty performing the training protocol even with frequent rest. Criteria for use of the arm support included weakness (2-/5 to 3-/5 muscle grades for elbow flexion and shoulder flexion, abduction, and rotation) and significant fatigue limiting functional performance as observed by the therapist. The subjects who met the criteria used the arm support for the duration of training at a set level of support. The level of support was customized for each subject by the therapist in order to optimize their ability to perform functional tasks and limit the impact of shoulder fatigue on grasp training.

Each participant completed 12 training sessions, three times per week for four weeks. Each training session was between 60-90 minutes including time for set up, system classification, donning/doffing the device, and rest breaks as needed. Participants completed 30 minutes of active training during each visit. After the 12 sessions, participants completed two days of post-testing, as described above, as well as a questionnaire for subjective feedback on their experience.

\subsection{Training}
\label{subsec:training}

The series of selected tasks reflect best-practice in UL prosthetic training (controls training, repetitive drills, and bimanual functional skills training)~\cite{johnson2014prosthetic}. During controls training, participants were educated on the device operation and demonstrated proficiency in the control absent any objects or functional tasks. Participants then advanced to repetitive drills, which incorporated an array of objects of various shapes, sizes, and densities (Fig.~\ref{fig:objects}).

Before donning the device, the therapist performed 5 minutes of passive range of motion to all joints of the UL to help mitigate fluctuations in tone across sessions. Then, the patient donned the device. During each session, participants completed 30 minutes of active training. Occasionally, breaks were provided upon patient request due to fatigue, or if any technical issues arose with the prototype device that required an adjustment be made during the session.

The training protocol was always carried out in the same order, with basic tasks first, progressing to more complex tasks. See Appendix~\ref{sec:training_procedure} for the list of tasks performed by subjects during each session. Some participants were not able to complete the full training protocol during each session. In that case, they stopped after 30 minutes of training and the last completed task was recorded. Some participants would complete the full protocol in less than 30 minutes. In that case, they continued working on grasp, transport, and release tasks of their choosing (to be client-centered) with oversight from the therapist for the duration of the 30 minutes.

\begin{table}
	\caption{Baseline characteristics}
	\centering
	\label{baseline_table}
	\begin{tabular}{c| c c c}
		\multirow{2}{*}{Assessment} & EMG group & SH group & All subjects\\
		& (n=6) & (n=5) & (n=11) \\ \hline \hline
		FM-Distal & 4.5 & 2 & 3.4 \\
		FM-Proximal & 23.2 & 20.6 & 22 \\
		FM-Total & 27.7 & 22.6 & 25.4 \\ \hline
		ARAT & 16.5 & 9.8 & 13.5 \\		
		BBT & 7.2 & 0.8 & 4.3
	\end{tabular}	
\end{table}

%\begin{table}[t!]
%	\caption{Subject demographics. Type of stroke : I(schemic), H(emorrogic), U(nknown).}
%	\hspace{-0.2cm}	
%	\label{demographics_table} 
%	\begin{tabular}{C{0.3cm}|C{0.25cm}C{0.25cm}C{0.5cm}C{0.4cm}C{0.7cm}C{0.9cm}C{1.05cm}C{0.85cm}}
%		& Age & Sex & Stroke type & Years since stroke & Affected side & Dominant hand & Intent detection & Arm support \\ \hline \hline
%		%%%%%%%%%% FM
%		s1  & 62 & F & I & 6 & L & R & EMG & N \\
%		s2  & 37 & F & I & 4 & L & R & EMG & N \\
%		s3  & 48 & F & I & 14 & R & R & EMG & N \\
%		s4  & 68 & M & U & 13 & R & R & EMG & N \\
%		s6  & 41 & M & I & 6 & R & R & SH & N \\
%		s7  & 32 & M & H & 22 & R & L & EMG & Y \\
%		s8  & 44 & M & I & 12 & R & R & SH & N \\
%		s9  & 44 & M & I & 5 & L & R & SH & N \\
%		s10 & 80 & F & I & 3 & L & R & EMG & N \\
%		s11 & 65 & F & I & 10 & R & R & SH & N \\
%		s12 & 50 & M & I & 2 & R & R & SH & N
%	\end{tabular}
%\end{table}

\section{Results}
\label{sec:results}

\begin{figure*}[t]
	\centering
		\includegraphics[width=0.49\linewidth]{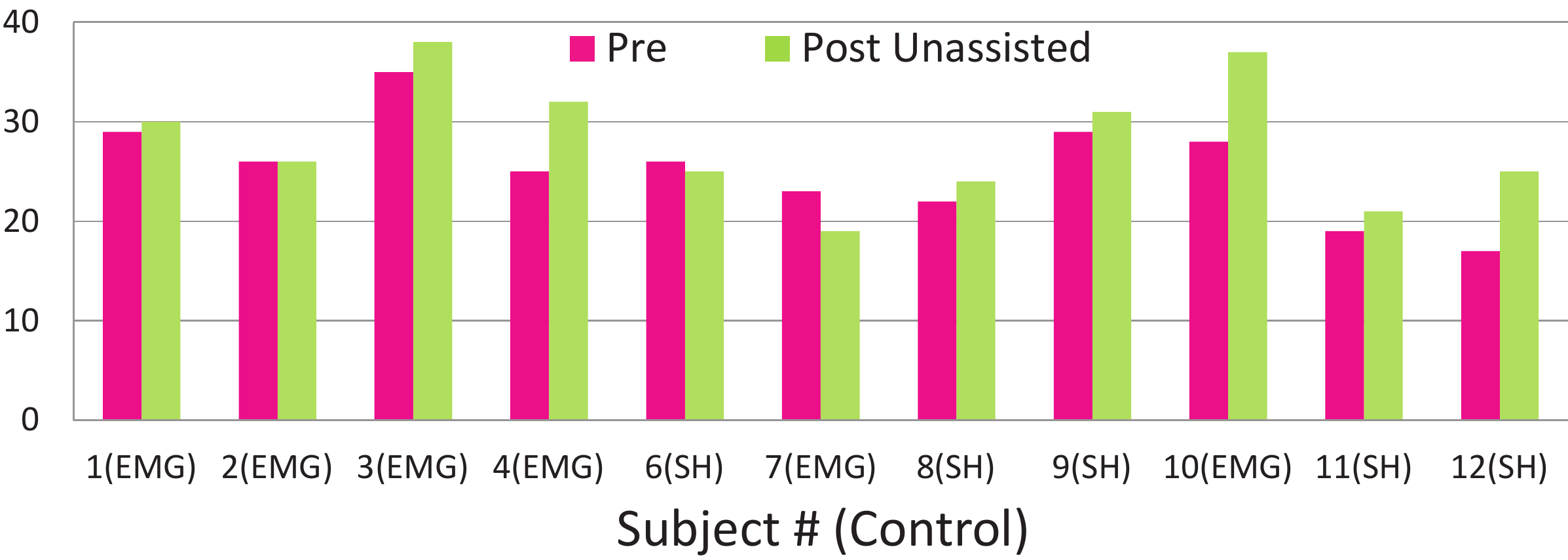} \hfill
		\includegraphics[width=0.49\linewidth]{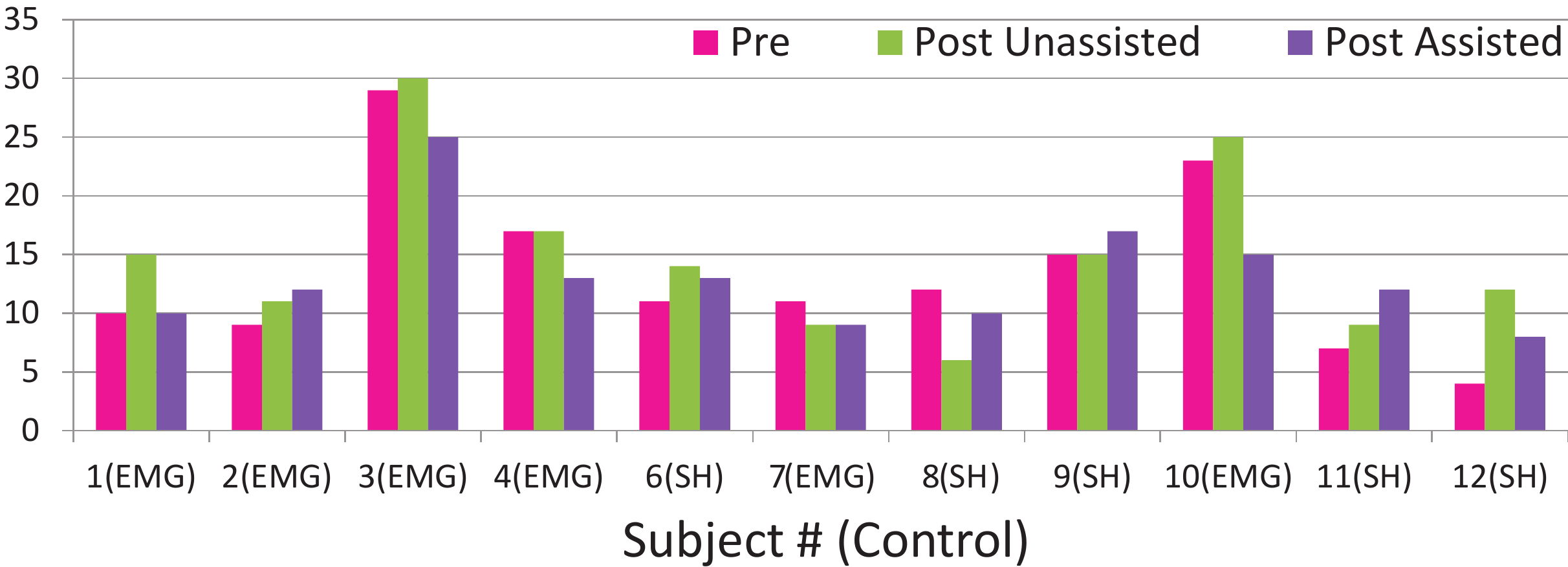}
	        \caption{FM(left) and ARAT(right) scores. Subject 5 dropped out due to a medical issue unrelated to the study.}
                \label{fig:total_results}                                
\end{figure*}

Among 12 enrolled individuals, 11 subjects (6 males and 5 females) completed the training and evaluations. One participant dropped out prior to the first training session due to a medical issue unrelated to the study, therefore all analysis is of 11 subjects. Years since stroke ranged from 2 to 22 years. 9 patients had ischemic, 1 had hemorrogic, and 1 had unknown type of stroke. There were 4 left affected and 7 right affected patients. 10 patients were right hand dominants and 1 was left hand dominant. 6 participants were screened to use EMG method and 5 subjects were assigned in SH group. Note that the subjects screened to the EMG group tended to have more functional use of their impaired UL at baseline, as noted by pre-testing scores compared to those assigned to the SH group (Table~\ref{baseline_table}). We examine the difference in clinical outcome between the two groups through secondary analyses. One subject used the arm support for training. 

\subsubsection{Fugl-Meyer Upper Extremity Scale}

The FM results are shown in Table~\ref{tab:FM}. We note that, at baseline, ten subjects had `no to poor' UL capacity ($<$31) and one subject had `limited capacity' (between 32 and 47) as defined in~\cite{hoonhorst2015fugl} on the FM. After the treatment, 8 subjects showed improvement, but 3 subjects did not; individual subject results are shown in Fig.~\ref{fig:total_results}. Overall, the participants showed a mean gain of 2.64 points without statistical significance. FM-distal (hand and wrist segments) improved significantly with a mean gain of 2.27 points, while significant improvement was not achieved in FM-proximal (shoulder and elbow segments). 86\% of the total mean gain (2.64 points) was attributed from the FM-distal. This is particularly notable given that the overall FM is more heavily weighted proximally, with more items in the FM-proximal (42/66) compared to FM-distal (24/66). 

\begin{table}[t]
	\caption{Gains for FM post-intervention. Bold data are statistically significant (P-value $<$ BH significance threshold).}
	\centering
	\label{tab:FM}
	\begin{tabular}{c |c c c}
		Category & Mean gain & P-value & Threshold \\ \hline \hline 
%		& (n=6) & (n=5) & (n=11)\\ \hline \hline
		%%%%%%%%%% FM
		Distal  & \textbf{2.27} & \textbf{0.001} & \textbf{0.006}\\
		Proximal & 0.36 & 0.372 & 0.042\\
		Total FM & 2.64 & 0.026 & 0.014
	\end{tabular}
\end{table}

\subsubsection{Action Research Arm Test}
ARAT results are shown in Table~\ref{tab:ARAT}. Mean gains were statistically significant in Grasp category when assisted post-training  was compared to baseline score, and when assisted post-training score was compared to post-training unassisted score. Post-training assisted performance was significantly worse when compared to post-training unassisted performance in the Grip category.

\subsubsection{Secondary Analyses}
Given the exploratory nature of this pilot study, we provide post-hoc secondary analyses to understand capabilities and limitations of the device in more detail and to highlight areas of focus for future work. Results for secondary analyses are shown in Table~\ref{tab:secondary_analyses}.

The FM results revealed that both the EMG (gain = 2.67, n = 6) and SH (gain = 2.6, n = 5) groups showed similar level of improvement while the EMG group showed more localized improvement (gain = 3, n = 6) in the distal segments than the SH group (gain = 1.4, n = 5).

When post-treatment ARAT was assessed without robotic assistance, positive mean gains of 1.33 points (n = 6) in the EMG group and 1.4 points (n = 5) in the SH group were achieved. For post-testing with robotic assistance, the EMG group performed worse by 2.5 points (n = 6, post-training assisted compared to pre-training) and 3.8 points (n = 6, post-training assisted compared to post-training unassisted) whereas the SH group showed mean gain of 2.2 points (n = 5, post-training assisted compared to pre-training) and 0.8 points (n = 5, post-training assisted compared to post-training unassisted). The most notable improvement in an ARAT subcategory was observed for the SH group in the Grasp category, with mean gain of 2.4 points (n = 5, post-training assisted compared to pre-training) and 3.0 points (n = 5, post-training assisted compared to post-training unassisted). Mean gains with robotic assistance in the SH group were negative in all other sub-categories.

For BBT, we broke down groups by level of functionality to examine outcome based on baseline functionality of the participants. At baseline, 7 participants scored 0 point, while 4 subjects scored between 4 to 23 points. When assisted by the robot the 7 participants without any baseline function achieved mean gains of 2 points (n = 7, post-training assisted compared to pre-training) and 1.71 points (n = 7, post-training assisted compared to post-training unassisted), but functional subjects performed worse both with and without the device.%The score drop in the EMG group with robotic assistance was likely because of poor performance of the EMG method when the user had to lift the arm higher due to the partition and height of the box. We note that the training sessions contained no action item that involved lifting an object higher than the height of the partition (15.2 cm), thus the participants did not get an opportunity to practice their proximal muscles in our protocol.

\subsubsection{Survey}

After post-testing, participants provided qualitative feedback through an open-ended survey. Participants generally reported enjoyment, functional improvements, and a desire to continue using the device. ``It encourages me to use my hand more. It gave me the feeling of freedom to use my hand again.'' ``I am able to fold and wring out a washcloth.'' ``If I could, I would wear it at home for most of the day for everything.'' Participants also offered feedback for device improvements such as reduced wiring, less bulk around fingertips, increased training intensity (time and duration), and actuation of the thumb tendon for powered pinch.

\begin{table}[t]
	\caption{Gains for ARAT post-intervention. Bold data are statistically significant (P-value $<$ threshold). (A):Post-training unassisted compared to pre-training, (B):Post-training assisted compared to pre-training, (C):Post-training assisted compared to post-training unassisted.}
	\centering
	\label{tab:ARAT}
	\begin{tabular}{c c|c c c}
		\multirow{2}{*}{Category}& Robotic & \multirow{2}{*}{Mean gain} & \multirow{2}{*}{P-value} & BH Significance \\ 
		&Assistance & & & Threshold \\ \hline \hline
		
		\multirow{3}{*}{Grasp} & (A) & 0.09 & 0.444 & 0.047 \\
		& (B) & \textbf{1.73} & \textbf{0.004} & \textbf{0.008} \\
		& (C) & \textbf{1.64} & \textbf{0.007} & \textbf{0.011} \\ \hline
		
		\multirow{3}{*}{Grip} & (A) & 0.82 & 0.054 & 0.017 \\
		& (B) & -0.64 & 0.176 & 0.036 \\
		& (C) & \textbf{-1.46} & \textbf{0.001} & \textbf{0.003} \\ \hline
		
		\multirow{3}{*}{Pinch} & (A) & -0.1 & 0.705 & 0.05 \\
		& (B) & -0.9  & 0.104 & 0.028 \\ 
		& (C) & -1 & 0.077 & 0.019 \\ \hline
		
		\multirow{3}{*}{\shortstack{Gross\\Movement}} & (A) & 0.37 & 0.271 & 0.039 \\
		& (B) & -0.54 & 0.161 & 0.033 \\
		& (C) & -0.91 & 0.079 & 0.022 \\ \hline

		\multirow{3}{*}{\shortstack{Total\\ARAT}} & (A) & 1.36 & 0.12 & 0.031 \\
		& (B) &  -0.36 & 0.385 & 0.044 \\
		& (C) & -1.72 & 0.103 & 0.025
	\end{tabular}
\end{table}

\begin{table} [t]
	\caption{Mean gains from FM, BBT and ARAT post-intervention for secondary analyses. (A), (B) and (C) are as in Table~\ref{tab:ARAT}.}
	\centering
        \vspace{-4mm}
	\subfloat[][FM Upper Extremity Scale]{
	\begin{tabular}{c|c c}
	  \multirow{2}{*}{Category} & EMG & SH \\
          & (n=5) & (n=5)\\ \hline \hline
		%%%%%%%%%% FM
		Distal & 3 & 1.4 \\ 
		Proximal& -0.33 & 1.2  \\
		Total & 2.67 & 2.6
	\end{tabular}
	\label{tab:FM_secondary}}
	%\qquad
	\subfloat[][Box and Block Test]{
	\begin{tabular}{c|c c c}
		Group & (A) & (B) & (C) \\ \hline \hline		
		%%%%%%%%%% BBT
		Functional & \multirow{2}{*}{-3.5} & \multirow{2}{*}{-9.25} & \multirow{2}{*}{-5.75}\\
                (n=4) & & & \\
		Non-func. & \multirow{2}{*}{0.29} & \multirow{2}{*}{2} & \multirow{2}{*}{1.71} \\
                (n=7) & & & 
	\end{tabular}
	\label{tab:BBT_secondary}}
	
	\subfloat[][Action Research Arm Test]{
	\begin{tabular}{c c|c c c}
		Category & Group & (A) & (B) & (C) \\ \hline \hline
		%%%%%%%%%% ARAT
		\multirow{2}{*}{Grasp} & EMG (n=6) & 0.67 & 1.17 & 0.5 \\ 
		& SH (n=5) & -0.6 & 2.4 & 3 \\ \hline
		\multirow{2}{*}{Grip} & EMG (n=6) & 0.33 & -1.8 & -2.17 \\ 
		& SH (n=5) & 1.4 & 0.8 & -0.6 \\ \hline 
		\multirow{2}{*}{Pinch} & EMG (n=6) & -0.17 & -1.83 & -1.67 \\ 
		& SH (n=5) & 0.4 & 0.2 & -0.2 \\ \hline
		\multirow{2}{*}{\shortstack{Gross\\Movement}} & EMG (n=6) & 0.5 & 0 & -0.5 \\ 
		& SH (n=5) & 0.2 & -1.2 & -1.4 \\ \hline
		\multirow{2}{*}{Total} & EMG (n=6) & 1.33 & -2.5 & -3.83 \\ %\cline{3-6}
		& SH (n=5) & 1.4 & 2.2 & 0.8
	\end{tabular}
	\label{tab:ARAT_secondary}}
	\qquad

	\label{tab:secondary_analyses}
\end{table}

\section{Discussion}
\label{sec:discussion}

Overall, we identified trends in the data that suggest this device might serve two distinct purposes for different subsets of the stroke population - namely as a rehabilitation or assistive device. But, the results also highlight limitations of the device, and point towards possible areas for future improvements.

\subsection{An Active Hand Orthosis for Rehabilitative Effects}

From a rehabilitation perspective, we discuss results obtained post-intervention without using the device, and compared to baseline performance. Positive gains noted on the FM-distal subtest suggest that training with the device can serve as a rehabilitative tool to remediate some functional use in the affected UL, especially to help improve hand functionality. In our secondary analyses, we note that the magnitude of gains on the FM-distal was larger in the EMG group than the SH group, suggesting that the restorative training effect may be greater in participants with some residual baseline functioning.

Based on the observation of a positive trend on the ARAT, we posit that increasing the intensity and duration of the intervention in future studies may lead to increased gains quantified using this measure. For example, small gains captured on the FM (e.g. ability to actively flex or extend the fingers) may not be captured on the ARAT because the improvement in range of motion was not sufficient to translate into increased functional ability (e.g. ability to pick up a small object).

\subsection{An Active Hand Orthosis for Assistive Effects}

We focus here on performance measured post-intervention with the participants actively using the device, and compare against pre-training baseline or post-training without robotic assistance. ARAT results suggest that, for stroke patients, using the robot as an assistive device for long term compensation to increase functionality in daily life may be feasible, although efficacy of the assistive capacity of this device has yet to be demonstrated.

\modifiedtext{AE-2}{The improvement in the Grasp category of the ARAT was expected as the device specifically assists with grasping tasks.} However, the magnitude of improvement was different depending on  control method, as seen in the secondary analyses. We speculate that the differences between EMG and SH groups may be related to baseline differences in hand functioning. EMG group participants tended to have more residual functioning at baseline, often employing compensatory patterns to achieve grasp and/or pinch, whereas when wearing the robotic device, the bulky finger components may have impaired their performance by making it more difficult to pick up and place objects in tight spaces. In contrast, those in the SH group had less functionality at baseline, and therefore the device assisted their ability to pick up objects, though they still had similar difficulty placing objects in tight spaces. 

%The most encouraging improvement in BBT was observed in participants with no functionality at baseline. We speculate that the negative gain by participants with non-zero baseline functionality occurred because these participants were employing functional compensatory pinch patterns at baseline. Based on our observations, however, training with the device discouraged compensatory patterns and forced users to grasp and pinch with typical patterns (e.g. finger pad to finger pad pinching), leading to poorer performance as participants did not master the new pinch pattern within the timeframe of the study. We believe additional training time may help, however, it is important to understand that some patients may be satisfied with their compensatory patterns if they are able to participate in their meaningful daily tasks. 

The outcome measures highlighted a few limitations of the device. Improvement in Grasp category of the ARAT and negative mean gains in all other categories implies that the device facilitated grasping of mid-size objects, but not small objects which require pinching. We speculate this was because the thumb was splinted into a stable opposition position against the other digits. This had the advantage of reducing actuator load, and we found this pose to be effective when grasping mid-size objects. However, a static thumb also made it difficult for subjects to stably hold small objects in a pinching pattern. This behavior likely affected the results in both the ARAT and BBT. As in recent studies, assisting the thumb has been shown effective in small object grasping~\cite{yurkewich2020hand,soekadar2016hybrid}, we are planning to address this issue in future studies by designing an actuated thumb component which enables assisted pinch in addition to enveloping larger objects.

Another limitation was poor performance of the EMG method due to abnormal muscle synergy in unregistered UL postures. The BBT score drop in the EMG group with robotic assistance was likely because of unstable EMG classification when the user had to lift the arm higher due to the partition and height of the box. We note that the training sessions contained no action item that involved lifting an object higher than the height of the partition (15.2 cm), thus the participants did not get an opportunity to practice their proximal muscles or learn to control the device while lifting the arm high in our protocol.

\addedtext{2-1}{The current design is not fully self-contained. Power is provided externally, and motor commands are sent from the computer over a wire. We are envisioning the next version containing a belt pack (weighing approximately 300 g) housing a battery pack and other electronics, including a microcontroller for issuing motor commands.}

\subsection{Limitations}
It is important to point out a number of limitations related to study design in this pilot case series. In particular, we did not use a control group consisting of patients receiving treatment of similar intensity and duration, but without robotic assistance. However, meaningful motor recovery with traditional physical therapy for chronic stroke patients with moderate to severe motor impairments is considered rare~\cite{duncan1992measurement,jung2002recovery}. In addition, our robotic-assistance enabled training tasks that were not possible for most participants otherwise, and thus can not be replicated with traditional therapy. Our study also did not comprise a follow up assessment to observe the durability of gains.  We plan to address these limitations in the future.

We also note that assignments between the EMG and SH groups was not performed in randomized fashion, but rather based on the ability of our intent inferral algorithm to classify EMG signals. As a result, we noticed systematic differences between the groups, with SH participants generally having lower baseline functionality. This limits our ability to interpret differences in the results obtained by the two groups. %\addedtext{AE-3}{We also note that our study performs post-hoc comparisons for secondary analyses which only include descriptive analysis. }

In addition, we note that rehabilitation studies strive to assess progress using outcome measures that are at the participation-level, observing and rating task performance in real-life environments. However, the FM measure, considered as the gold-standard in stroke research due to well established psychometric properties and MCID~\cite{page2012clinically}, only assesses body structures at the impairment level, focusing on the capacity of the UL to move. ARAT and BBT are activity-level assessments that involve observing and scoring participants performing simulated functional tasks that are shorter in duration and more highly scripted compared to ADLs. We believe customized outcome measures that capture higher task variation and allow longer completion times might be better suited for capturing progress when using robotic devices in an assistive fashion.

While it is important for patients to be able to don and doff a wearable device without assistance, we are still far from quantifying this characteristic. With the current prototype, supervision is required as it would be for traditional therapy, but in the future as the device is further refined, it is hoped that patients will be able to use the device independently at home after initial training with a therapist.

Finally, we did not conduct any structured interviews or ask the users to rate the device using standard usability scales. However, participants were asked to provide open-ended, qualitative feedback. Furthermore, a trained occupational therapist supervised each training session and  thoroughly monitored for pressure, redness, pain, and any other signs of distress throughout the study. There have been no complaints from patients reported in this study. As the system is further refined, we plan to use standard questionnaires, such as System Usability Scale (SUS) or Likert scale.

\section{Conclusion}
\label{sec:conclusion}

In this work, we presented clinical outcomes after 12 training sessions for a study using a robotic hand orthosis. Our device is designed to assist the paretic hand after stroke, focusing primarily on an impairment pattern characterized by difficulty with active finger extension. Two main design goals for our device are wearability and user-driven operation: we use two different methods to infer the intent of the user, and thus to determine when to provide assistance. %Participants were evaluated with the FM (unassisted only) and the ARAT and BBT (both assisted and unassisted) to study the utility of the robot as a rehabilitative training device and/or an assistive device .

%The first method (EMG) relies on ipsilateral surface electromyographic signals, while the second one (SH) uses contralateral shoulder movement detected via an instrumented shoulder harness.

%A wearable, user-driven hand orthosis could fulfill two roles: a rehabilitation device, designed to improve performance of the unassisted UL by providing assisted exercise, or an assistive device, designed to improve performance of the assisted UL. In this work, we have presented the clinical outcomes of a clinical intervention designed as an early feasibility test of both of these hypotheses. In our study, eleven chronic stroke participants engaged in a month-long training protocol in which they trained with our orthosis. The subjects were assigned to either an EMG or an SH group for intent inferral based on a control screening, and they all used the exotendon device which assisted with finger extension for functional grasp activities during training. 

Post-intervention FM sub-scores suggest the grasp exercises helped improve distal movements of UL whereas it did not have a significant impact on proximal segments. This result suggests the possibility for using our orthosis as a rehabilitative device for the hand. Assisted ARAT scores show that the device can indeed function in an assistive role for stroke patients. However, the results should be cautiously interpreted because of the limited sample size, \addedtext{AE-4}{lack of a control group}, and the fact that the outcomes did not meet MCID for either FM or ARAT. We are planning to address this aspect through both improvements to the device and extended training periods in future work.%We noted no significant differences between the EMG and SH groups. This is, to the best of our knowledge, the first time that either assistive or rehabilitative effects were shown via clinical outcome measures by using a wearable, fully user-driven robotic hand orthosis.

Our study also underscored limitations of the device. In particular, the device disrupts compensatory grasp patterns developed by stroke survivors, leading to an immediate decrease in functionality when the device is removed. It is likely that the 12 sessions were not long enough to enable learning of new grasp patterns for participants. Furthermore, our current design relies on a static, passively splinted thumb, which enables gross grasp but is not suited for pinching smaller objects. %Our clinical intervention did not include a control group receiving traditional therapy instead of robotic assistance, or a follow-up assessment.

Nevertheless, we believe that this work can highlight the potential and feasibility for wearable and user-driven robotic hand orthoses. Such devices may enable robotic based-hand rehabilitation during daily activities (as opposed to isolated hand exercises with limited UL engagement) and over extended periods of time, even in a patient's home environment. Numerous challenges must still be overcome in order to achieve this vision, related to design (compact devices with easier donning/doffing), control (robust yet intuitive intent inferral), and effectiveness (improved functionality in a wider range of metrics). However, if these challenges can be addressed, wearable robotic devices have the potential to greatly extend the use and training of the affected UL after stroke, and help improve the quality of life for a large patient population.
\begin{appendices}
\section{Training Procedure}
\label{sec:training_procedure}

Participants picked up and released selected objects 5 times with their forearm resting on the table top (supported reach) and then 5 times with their forearm lifted off the table (unsupported reach) to simulate functional reach. The objects used in this task include (1) 2.5 cm wooden cube, (2) 5 cm wooden cube, (3) tennis ball, (4) 4 cm diameter toiletry bottle, and (5) 13 cm tall, tapered, hard plastic cup. Objects were positioned in various locations to optimize the functional envelope. The objects were then arranged on a tray with 1 inch raised lip and participants removed all 5 items from the tray twice, and then replaced all items onto the tray twice.

Next, participants picked up and released the following irregularly shaped objects twice each: (1) cotton ball, (2) 1 inch rubber ball, and (3) washcloth.

Lastly, participants advanced to bimanual functional skills training to best simulate real-life conditions and the additional physical and cognitive challenge of operating the device without exclusive attention to its performance. The tasks were completed twice each with participants using their affected hand to stabilize and their unaffected hand to perform manipulation: (1) removing and replacing the cap from a broad line marker, (2) removing and replacing the cap from a standard tube of toothpaste (screw off), (3) removing and replacing the cap from a screw top beverage bottle, (4) removing and replacing the wide mouth screw cap from large coffee container, (5) using a wooden spoon to stir in a small bowl for 10 seconds (affected hand stabilized bowl), (6) using a butter knife to make 2 cuts in a `log' of theraputty to mimic cutting food (affected hand stabilized theraputty), (7) opening a lock with a key (affected hand held lock), and (8) opening a sealed sandwich-size ziploc bag.
\end{appendices}

\bibliographystyle{IEEEtran}
\bibliography{bib/orthoses,bib/stroke}     % need at least one citation not to cause an error

\end{document}